\documentclass[conference]{IEEEtran}
\IEEEoverridecommandlockouts
\usepackage{cite}
\usepackage{amsmath,amssymb,amsfonts}
\usepackage{algorithmic}
\usepackage{graphicx}
\usepackage{textcomp}
\usepackage{xcolor}
\usepackage{booktabs}
\usepackage[utf8]{inputenc} 
\usepackage[T1]{fontenc}    
\usepackage{url}            
\usepackage{multirow}
\usepackage{nicefrac}       
\usepackage{microtype}      
\usepackage{tabularx}
\usepackage{verbatim }
\usepackage{makecell}

\def\BibTeX{{\rm B\kern-.05em{\sc i\kern-.025em b}\kern-.08em
    T\kern-.1667em\lower.7ex\hbox{E}\kern-.125emX}}
    
\begin{document}

\title{IPixMatch: Boost Semi-supervised Semantic Segmentation with Inter-Pixel Relation\\
}



\author{\IEEEauthorblockN{Kebin Wu}
\IEEEauthorblockA{
\textit{Technology Innovation Institute}\\
Abu Dhabi, United Arab Emirates \\
kebin.wu@tii.ae}
\and
\IEEEauthorblockN{Wenbin Li}
\IEEEauthorblockA{
\textit{Technology Innovation Institute}\\
Abu Dhabi, United Arab Emirates \\
wenbin.li@tii.ae}
\and
\IEEEauthorblockN{Xiaofei Xiao}
\IEEEauthorblockA{
\textit{Technology Innovation Institute}\\
Abu Dhabi, United Arab Emirates \\
xiaofei.xiao@tii.ae}
}

\maketitle

\begin{abstract}
The scarcity of labeled data in real-world scenarios is a critical bottleneck of deep learning's effectiveness. Semi-supervised semantic segmentation has been a typical solution to achieve a desirable tradeoff between annotation cost and segmentation performance. However, previous approaches, whether based on consistency regularization or self-training, tend to neglect the contextual knowledge embedded within inter-pixel relations. This negligence leads to suboptimal performance and limited generalization. In this paper, we propose a novel approach \textbf{IPixMatch} designed to mine the neglected but valuable \textbf{I}nter-\textbf{Pix}el information for semi-supervised learning. Specifically, IPixMatch is constructed as an extension of the standard teacher-student network, incorporating additional loss terms to capture inter-pixel relations. It shines in low-data regimes by efficiently leveraging the limited labeled data and extracting maximum utility from the available unlabeled data. Furthermore, IPixMatch can be integrated seamlessly into most teacher-student frameworks without the need of model modification or adding additional components. Our straightforward IPixMatch method demonstrates consistent performance improvements across various benchmark datasets under different partitioning protocols.
\end{abstract}


\section{Introduction}
\label{sec:intro}
Semantic segmentation is a pixel-level classification task, with wide-range of applications in domains like autonomous driving and heath care. Present studies predominantly rely on supervised learning, for which numerous pixel-wise annotated data are required. However, the process of data labeling for segmentation is costly and time-consuming. To alleviate the annotation burden and to tackle the challenge of learning from small labeled datasets, semi-supervised semantic segmentation has aroused gradually-growing interest by leveraging a limited number of pixel-level labeled images and a large amount of unlabeled data, both of which are drawn from the same distribution. The key challenge within semi-supervised learning lies in how to effectively utilize the unlabeled data, so that more valuable information of the data distribution can be extracted. 

Existing efforts in the field of semi-supervised segmentation can typically be grouped into two categories: self-training and consistency regularization. Self-training methods assign pixel-wise pseudo-labels to unlabeled data based on the model's predictions. These pseudo-labeled data are then leveraged together with labeled data in the model training. By iteratively alternating between pseudo-labeling and model training, self-training manages to mine knowledge from the unlabeled data, thereby achieving competitive performance. Since the quality of the predicted pseudo-labels heavily affects the semi-supervised performance, various algorithms have been proposed to mitigate the influence of unsatisfying pseudo-labels. This includes techniques such as setting thresholds for positive samples~\cite{yang2023revisiting} and negative samples~\cite{wang2022semi}, explicit handling of class distribution~\cite{he2021re}, and addressing class imbalance in pseudo-labeling~\cite{guan2022unbiased}. As for consistency regularization, it designs a pretext task for unlabeled data by imposing different pixel-level perturbations on the images and then aligning their output predictions. Recent studies have focused on designing diverse perturbations, including augmentations applied to image~\cite{yuan2021simple}, noise injected to feature space~\cite{ouali2020semi}, as well as perturbation in the network such as using different initializations~\cite{ke2020guided}. 

While these approaches demonstrate certain levels of effectiveness, both self-training and consistency regularization suffer from some weaknesses. On one hand, most prior works in self-training adopt one-hot hard pseudo-labels, overlooking hidden information present in predictions. On the other hand, the consistency in previous studies are often considered in a pixel-wise manner, lacking the capability to capture spatial dependency. Consequently, the resulting model prediction tends to be less context-aware of spatial relations. In order to address the weaknesses, we propose a novel semi-supervised semantic segmentation paradigm named \textbf{IPixMatch}. It captures the latent \textbf{I}nter-\textbf{Pix}el dependencies embedded within soft pseudo-labels by minimizing a specifically designed loss term, allowing to boost the generalization ability of the resulted model. 
Such design enables IPixMatch to be applicable to most semi-supervised segmentation approaches. 
Under most of the evaluation protocols, our remarkably straightforward approach achieves improvements. Moreover, our approach demonstrates an effective strategy for learning from small data, a common and challenging scenario in many practical applications, by maximizing the informational yield from every labeled instance and intelligently exploiting the structure within unlabeled data.

In summary, we have the following contributions.

\begin{itemize}
\item We point out that the inter-pixel relation in pseudo-labels are not fully investigated in the existing semi-supervised semantic segmentation. 
\item We propose a novel semi-supervised learning paradigm IPixMatch, which fully probes the inter-pixel dependencies inherent in the soft pseudo-labels, allowing to boost the generalization ability of the model.
\item Remarkably straightforward, our approach achieves improvements under most of the evaluation protocols. 
\end{itemize}

\section{Related Work}

Semi-Supervised learning (SemSL) trains a model using a limited number of labeled data and abundant unlabeled data. 
Typically, there are two methodologies for semi-supervised learning, namely consistency regularization~\cite{berthelot2019mixmatch,xie2020unsupervised,laine2017temporal} and self-training~\cite{xie2020self,lee2013pseudo,grandvalet2004semi}. Consistency regularization aims to optimize the model by enforcing consistent predictions for images under various perturbations. Some works also explore the consistency constraint, when the perturbations come from the feature space or the network itself. In contrast, self-training is designed to assign pseudo-labels to unlabeled data, which are then used together with the labeled data for training in an iterative manner. FixMatch~\cite{sohn2020fixmatch}, which is an elegant fusion of these two methods, proposes to impose strong perturbations on the unlabeled images, while borrows pseudo-supervisions from predictions of the weakly perturbed ones at the same time. 


Recent advancements in semi-supervised semantic segmentation have followed a similar trajectory to SemSL. The work in the line of self-training implements pixel-wise pseudo-labeling to get auxiliary supervision for unlabeled images~\cite{chen2021semi,hu2021semi,wang2022semi}, while algorithms based on consistency assumption aim to enforce the model's pixel-wise prediction to be invariant to different forms of perturbations~\cite{lai2021semi,zhong2021pixel,zhang2022region}. In the spirit of FixMatch in SemSL, consistency regularization and pseudo-labeling are cooperated harmoniously in the field of semantic segmentation as well. Impressive success it is, such combination has not fully overcome the weaknesses of consistency prior and pseudo-labeling (e.g., proper consistency terms, or noisy pseudo-label). 

Therefore, on top of the framework of FixMatch and its variants, most recent approaches in semi-supervised semantic segmentation explore improvements by designing more intensive and reasonable consistency terms, as well as by mitigating the negative influence of noisy pseudo-labels. UniMatch~\cite{yang2023revisiting} adapts FixMatch to pixel level, and explores deeper and wider augmentations by introducing perturbation in feature space and incorporating a dual-stream of strong perturbations in the image level. While such designs with pixel-wise augmentations for consistency yield substantial performance improvements, the enforced consistencies are limited in the pixel level. We highlight that inter-pixel relations convey valuable information, which however have not been explored in previous semi-supervised segmentation. 

\section{Method}

In Semi-Supervised Semantic Segmentation, we have a labeled dataset $\mathcal{D}_L = \{(\mathbf{x_i}, \mathbf{y_i})\}^{|\mathcal{D}_L|}_{i=0}$, where $\mathbf{x_i}$ is the input image with a spatial size of $H \times W $, and $\mathbf{y_i}$ is the ground truth segmentation map with $C$ semantic classes. In addition, another unlabeled dataset $\mathcal{D}_U = \{(\mathbf{x_i^u})\}^{|\mathcal{D}_U|}_{i=1}$ is given, with a common assumption being that $|\mathcal{D}_L| \ll |\mathcal{D}_U|$. The goal of semi-supervised semantic segmentation is to maximize the utility of the limited labeled data and the substantial volume of unlabeled data to achieve high segmentation performance.

\subsection{Neglected inter-pixel relation in traditional pseudo-labeling}

Pseudo-labeling assigns pseudo-labels for unlabeled samples based on the model predictions. Denoting the output of a model $\mathcal{M}$ for a given image $\mathbf{x^u}$ as $\mathbf{p} = \mathcal{M}(\mathbf{x^u})$,  then the traditional hard pseudo-labeling can be represented as:
\begin{equation}\label{eq:pseudolabeling}
  \mathbf{\widehat{y}}^{j} = \arg \underset{c}{\max}{\ \mathbf{p}^{j}_c},
\end{equation}
where $c = 1, 2, ..., C$ indexes the semantic class, $j = 1, 2, ..., H \times W $ means the $j^{th}$ pixel in the segmentation map, and  $\mathbf{\widehat{y}}^{j}$ represents its pseudo-class. 

Given a model's prediction output, most existing solutions adopt Eq. \ref{eq:pseudolabeling} to assign pseudo-labels. In this paper, we term such approach as hard pseudo-labeling thereafter. Popular it is, hard pseudo-labeling has two drawbacks.

Firstly, for each pixel, its prediction $\mathbf{p}^j$ conveys the confidence of all semantic classes. However, hard pseudo-labeling simply selects the index corresponding to the maximum value, discarding the relative probabilities associated with different semantic classes.

Secondly, the inter-pixel relation is not considered. For a given channel (\textit{e.g.}, $c^{th}$) in the prediction $\mathbf{p}$, it encodes the saliency knowledge for the corresponding (\textit{i.e.}, $c^{th}$) semantic class~\cite{shu2021channel}. The adoption of hard pseudo-labeling leads to the generation of binary, category-specific saliency maps, inadvertently undermining the potential of inter-pixel relation. We argue that the inter-pixel relation brings extra insight to guide the semi-supervised learning. For instance, the inter-pixel knowledge can reveal which region(s) to be attended for a certain scene category.

In light of these observations, we propose IPixMatch, a novel framework crafted to harness the latent inter-pixel relation embedded within soft (instead of hard) pseudo-labels for semi-supervised learning. 

\subsection{IPixMatch}
\label{sec: method-darkmatch}

\begin{figure*}[t]
  \centering
   \includegraphics[width=0.6\linewidth]{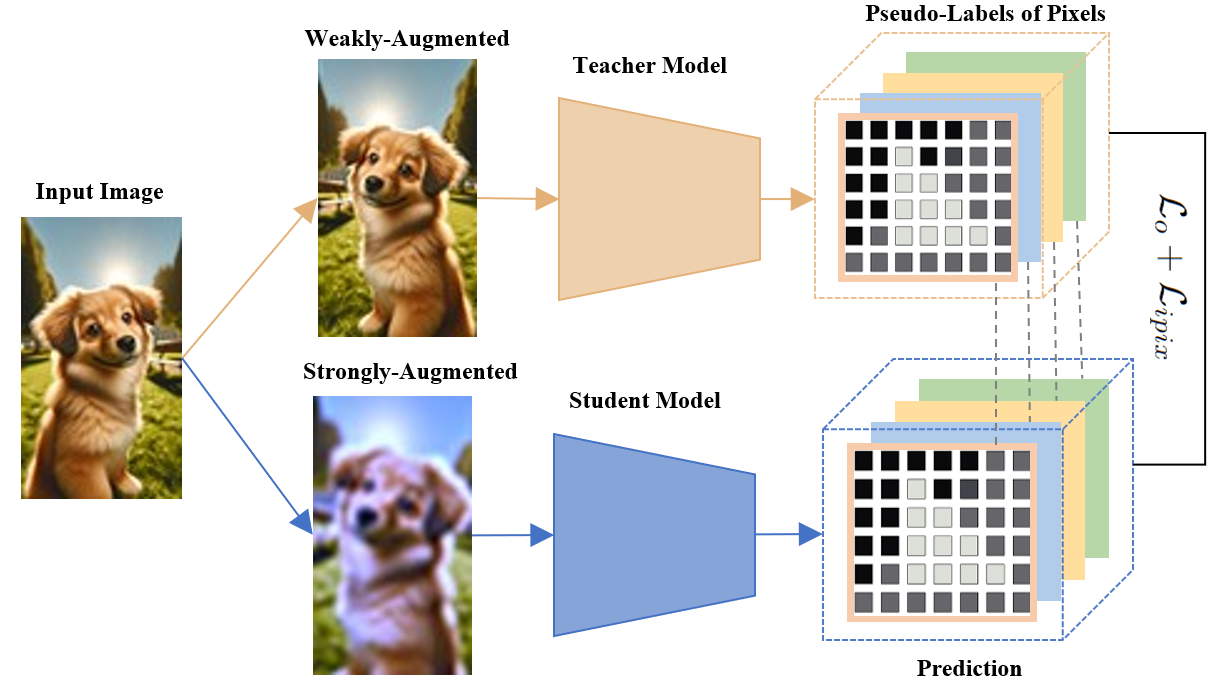}   \caption{IPixMatch framework to mine the inter-pixel relation, where each line denotes spatial relation consistency mapping per channel.}
   \label{fig:interspatial}
\end{figure*}

Fig.~\ref{fig:interspatial} illustrates the flowchart of IPixMatch. In IPixMatch, the well-established teacher-student structure \cite{sohn2020fixmatch} is adopted. Within this framework, the teacher model is responsible for pseudo-labeling which takes weakly augmented images as input, whereas the student model leverages strongly augmented images for the model optimization under the guidance of the original semi-supervised segmentation loss $\mathcal{L}_{o}$ and the inter-pixel loss $\mathcal{L}_{IPix}$, which are detailed later in this section. As with most setting, the teacher and student models share identical architectures, and the parameters of the teacher are updated with an exponentially moving average (EMA) of the weights of the student. In the semi-supervised segmentation, IPixMatch alternates between pseudo-labeling and model training in an iterative manner. 

Given an unlabeled image, assume the logit outputs from the teacher model and student model are $\mathbf{p} \in \mathcal{R}^{C \times {HW}}$ and $\mathbf{q} \in \mathcal{R}^{C \times {HW}}$, respectively. To capture the spatial relation conveyed in $\mathbf{p}$ and then enforce the student model to learn such relation, an inter-pixel loss $\mathcal{L}$ is defined in Eq.~\ref{eq:interspatial_loss}. 

\begin{gather}\label{eq:interspatial_loss}
  \mathcal{L} = \frac{1}{C\cdot{\log(|\Omega|)}} \sum_{k=1}^{C} \phi(\psi({\mathbf{p}_{k,\Omega}}), \psi({\mathbf{q}_{k,\Omega}})) \\
  \Omega = \{\,j \mid \mathrm{max}( {\operatorname{Softmax}(\mathbf{p}_{:,j}))} >{\tau}\,\},
\end{gather}

\noindent where $\Omega$ is the set of pixels whose largest class confidence exceeds a predefined threshold ${\tau}$ (set to 0.8 by default). $\psi$ is a function that converts logits to probability distributions, and $\phi$ calculates the discrepancy distance between $\psi({\mathbf{p}_{k,\Omega}})$ and $\psi({\mathbf{q}_{k,\Omega}})$. 

Typically, a common choice to measure the distance between two distributions is to leverage Kullback–Leibler (KL) divergence~\cite{kullback1951information}. Besides, to convert logits to probability distributions, the following function is adopted.
\begin{equation}\label{eq:softmax}
  \psi(\mathbf{z}_i) = \frac{e^{\mathbf{z}_{i}/t}}{\sum_{j=1}^{HW} e^{\mathbf{z}_{j}/t}} \ \ \ for\ i=1,2,\dots,HW
\end{equation}
where $t$ controls the smoothness degree of the targeted distribution and $\mathbf{z}$ is a vector of logits with a length of $HW$. In this paper, $t$ is set to 4 by default. When training goes on, more pixels are gradually included in $\Omega$, which in turn makes the KL distance fluctuate. To mitigate these fluctuations, the KL distance is normalized by the term $\log(|\Omega|)$. The resulting loss function, formulated in this manner, is denoted as $\mathcal{L}_{KL}$.

While KL distance proves effective in some scenarios, it was highlighted in~\cite{huang2022knowledge} that such KL based consistency regularisation may be harmful if the divergence between the two distributions is severe. Therefore, a correlation-based metric was proposed in~\cite{huang2022knowledge} to measure the distribution distance, which aims to relax the demanding matching in KL by explicitly focusing on relative relations in the distribution vector only. The distance $\phi$ is defined as follows
\begin{equation}\label{eq:dist_loss}
  \phi(\mathbf{u},\mathbf{v}) =  1 - \rho_{p}(\mathbf{u},\mathbf{v}),
\end{equation}
where $\mathbf{u}$ and $\mathbf{v}$ are two distribution vectors, and $\rho_p$ is the Pearson correlation coefficient. After adopting such correlation-based distance measurement in Eq.~\ref{eq:interspatial_loss}, our inter-pixel consistency loss $\mathcal{L}$ will be referred to as $\mathcal{L}_{CR}$ thereafter. By minimizing this consistency loss term, which can be either KL based $\mathcal{L}_{KL}$ or the correlation-based $\mathcal{L}_{CR}$, the inter-pixel relation is harnessed for semi-supervised segmentation.

To make sure IPixMatch can be seamlessly integrated with existing semi-supervised segmentation methods, we adopt a straightforward yet effective method by adding a weighted loss term $\mathcal{L}_{IPix}$ to the original loss function $\mathcal{L}_{o}$ in a specific semi-supervised segmentation approach, as shown below.

\begin{equation}\label{eq:final_loss}
\mathcal{L}_{sum} = \mathcal{L}_{o}  + \alpha \cdot \mathcal{L}_{IPix}
\end{equation}
where $\alpha$ is a balancing weight and $\mathcal{L}_{IPix}$ can be either $\mathcal{L}_{KL}$ or $\mathcal{L}_{CR}$. $\mathcal{L}_{o}$ is the sum of loss terms used in a existing semi-supervised segmentation algorithm, which are defined according to SemSL strategy such as consistency training \cite{yang2023revisiting,ouali2020semi} or self-training \cite{wang2022semi,yang2022st++}.

\begin{table*}[t!]
\caption{Effects of different loss terms on PASCAL VOC 2012 under two different protocols. UniMatch* is added to serve as baselines. The average improvements over the corresponding baselines are presented at the bottom. \textbf{Bold} value indicates the best.}
\label{table:loss term}
  \begin{center}
  \newcolumntype{Y}{>{\raggedleft\arraybackslash}X}
    \begin{tabularx}{\textwidth}{X|XXX} 
     \toprule
        Labeled Num & UniMatch* & $\mathcal{L}_{KL}$ & $\mathcal{L}_{CR}$ \\   
      \midrule 
      \#P1: 1/16 (662)  & {73.33} &	{74.68} & {74.22} \\
      \#P2: 92 \quad \quad   &  {70.12}  & {70.36} &  {69.61}  \\
      \midrule 
      Average $\uparrow$ & {-}  & \textbf{0.80} & {0.19} \\   
\midrule 
\bottomrule 
\end{tabularx}

\end{center}
\end{table*}

\begin{table*}[t!]
\caption{IPixMatch with and without warmup. \textbf{Bold} values indicate the best, while \textit{italic} values represent the mIOU improvements after adopting warmup.}
\label{table:warmup}
  \begin{center}
  \newcolumntype{Y}{>{\raggedleft\arraybackslash}X}
    \begin{tabularx}{\textwidth}{l|XXX|XXX} 
     \toprule
      \multirow{2}{*}{ \textbf{PASCAL VOC 2012} } & \multicolumn{3}{c|}{ ResNet-50 } & \multicolumn{3}{c}{ ResNet-101 } \\
      \cmidrule{2-7}  & $1/16$  & $1 / 8$ & $1 / 4$ & $1 / 16$ & $1 / 8$ & $1 / 4$ \\
      \midrule 
      IPixMatch | without warmup & {75.36} &	\textbf{77.07}  & {76.67}  & {77.12} & {78.14} & {78.78} \\
      IPixMatch | with warmup &  \textbf{75.69} & {76.66}  & \textbf{77.16}  & 
      \textbf{77.16} & \textbf{78.15} & \textbf{78.83} \\
        \textit{Improvement by warmup} & \textit{+0.33} &	\textit{-0.41}  & \textit{+0.49}  & \textit{+0.04} & \textit{+0.01}  & \textit {+0.05}\\			
\midrule 
\bottomrule 
\end{tabularx}
\end{center}
\end{table*}

\section{Experiments}
In this section, we detail the datasets used, outline implementation specifics, and provide a comprehensive study to validate the effectiveness of IPixMatch.

\subsection{Dataset}
\textbf{PASCAL VOC 2012} originally consists of 1,464 fully annotated high-quality images for training and 1449 images for validation~\cite{everingham2015pascal}. Later, it is augmented using SBD~\cite{hariharan2011semantic}, which contains low-quality and noise contaminated labeled images, thereby extending the number of the labeled samples to 10,582. To validate the semi-supervised segmentation, two types of protocols are often utilized in previous work to sample labeled data: (1) from the extended training images (denoted as \#P1 thereafter) and (2) from the original high-quality training images (denoted as \#P2 thereafter). 

\textbf{Cityscapes} is another benchmark segmentation dataset and its images are captured on urban scenes~\cite{cordts2016cityscapes}. In total, the dataset has 2,975 annotated high-resolution training images and an additional 500 validation images. 

\begin{table*}[t!]
\caption{Comparison with SOTA methods on PASCAL VOC 2012 under protocol \#P1. The fractions 1/16, 1/8, and 1/4 represent the proportion of labeled data. UniMatch* denotes our reproduced results of UniMatch, following the modifications mentioned in Sec.~\ref{sec: exp.detail}. The numbers 321 and 513 are image resolution used in training. \textbf{Bold} indicate the best, \textit{italics} indicates the mIOU improvement of IPixMatch against UniMatch*.}
\label{table:pascal-P1}
  \begin{center}
  \newcolumntype{Y}{>{\raggedleft\arraybackslash}X}
    \begin{tabularx}{\textwidth}{l|XXX|XXX} 
     \toprule
      \multirow{2}{*}{ \textbf{PASCAL VOC 2012} } & \multicolumn{3}{c|}{ ResNet-50 } & \multicolumn{3}{c}{ ResNet-101 } \\
      \cmidrule{2-7}  & $1/16$  & $1 / 8$ & $1 / 4$ & $1 / 16$ & $1 / 8$ & $1 / 4$ \\
      \midrule 
      Supervised Baseline & 61.2 & 67.3 & 70.8 & 65.6 & 70.4 & 72.8 \\
      CAC \cite{lai2021semi} & 70.1 & 72.4 & 74.0 & 72.4 & 74.6 & 76.3 \\
      \midrule 
      UniMatch*~\cite{yang2023revisiting}  \ \ \,| \textcolor{darkgray}{\textbf{321}} & 73.3	& 74.8	& 74.5	& 75.2	& 75.8	& 76.5 \\
      {IPixMatch (ours)} | \textcolor{darkgray}{\textbf{321}} & \textbf{74.5} &	\textbf{74.9}  & \textbf{75.0}  & \textbf{75.9} & \textbf{76.2} & \textbf{77.1} \\  
      \textit{Improvement} & \textit{+1.2} &	\textit{+0.1}  & \textit{+0.4}  & \textit{+0.7} & \textit{+0.4} & \textit{+0.6} \\
      \midrule 
      \midrule 
    Supervised Baseline & 62.4 & 68.2 & 72.3 & 67.5 & 71.1 & 74.2 \\
      CPS~\cite{chen2021semi} & 72.0 & 73.7 & 74.9 & 74.5 & 76.4 & 77.7 \\
  U$^2$PL~\cite{wang2022semi} & 72.0 & 75.1 & 76.2 & 74.4 & 77.6 & 78.7 \\
 PS-MT~\cite{liu2022perturbed} & 72.8 & 75.7 & 76.4 & 75.5 & 78.2 & 78.7 \\
       \midrule 
 UniMatch*~\cite{yang2023revisiting} \ \ \,| \textcolor{darkgray}{\textbf{513}} & 75.6 & 76.5 & 76.9 & 76.6 & 77.7 & 78.5 \\
 {IPixMatch (ours)} | \textcolor{darkgray}{\textbf{513}} & \textbf{75.7} & \textbf{76.7} & \textbf{77.2} & \textbf{77.2}	& \textbf{78.2}	& \textbf{78.8} \\ 
      \textit{Improvement} & \textit{+0.1} &	\textit{+0.2}  & \textit{+0.3}  & \textit{+0.6} & \textit{+0.5} & \textit{+0.4} \\ 
      \midrule 
      \bottomrule 
\end{tabularx}
\end{center}
\end{table*}

\subsection{Implementation details}
\label{sec: exp.detail}

\begin{table*}[t!]
 \caption{Comparison with SOTA methods on PASCAL VOC 2012 under protocol \#P2. The numbers in the head 92, 183, 366, 732, and 1464 represent the number of labeled data. UniMatch* denotes our reproduced results of UniMatch, following the modifications in Sec.~\ref{sec: exp.detail}. \textbf{Bold} indicates the best, \underline{underline} represents the second best, and \textit{italic} values report the mIOU improvement of IPixMatch over UniMatch. Following \cite{yang2023revisiting}, results for IPixMatch and Unimatch here utilize the image size of 321, while the others adopt 512.}
\label{table:pascal-P2} 
  \begin{center}
  \newcolumntype{Y}{>{\raggedleft\arraybackslash}X}
    \begin{tabularx}{\textwidth}{l|XXXXX} 
     \toprule
      { \textbf{PASCAL VOC 2012} } & 92 & 183 & 366 & 732 & 1464 \\
      \midrule 
      Supervised Baseline & 44.0 & 52.3 & 61.7 & 66.7 & 72.9 \\
      PC$^2$Seg~\cite{zhou2021c3} & 56.9 & 64.6 & 67.6 & 70.9 & 72.3 \\
      \midrule 
      UniMatch*~\cite{yang2023revisiting} \ \ \,| \textcolor{darkgray}{\textbf{RN-50 }} & \underline{70.1}	& \underline{71.7}	& \underline{74.6}	& \textbf{76.4}		& \underline{77.7} \\
      {IPixMatch (ours)} | \textcolor{darkgray}{\textbf{RN-50 }}  & \textbf{70.8} &	\textbf{72.4}  & \textbf{75.1}  & \underline{76.0} & \textbf{78.1}  \\    
      \textit{Improvement} & \textit{+0.7} &	\textit{+0.7}  & \textit{+0.5}  & \textit{-0.4} & \textit{+0.5} \\      
      \midrule 
      \midrule 
      Supervised Baseline & 45.1 & 55.3 & 64.8 & 69.7 & 73.5 \\
     CPS~\cite{chen2021semi} & 64.1 & 67.4 & 71.7 & 75.9 & - \\
     ST++~\cite{yang2022st++}  & 65.2 & 71.0 & 74.6 & 77.3 & 79.1 \\
     U$^2$PL~\cite{wang2022semi} & 68.0 & 69.2 & 73.7 & 76.2 & \underline{79.5} \\
     PS-MT~\cite{liu2022perturbed} & 65.8 & 69.6 & 76.6 & 78.4 & \textbf{80.0} \\
     \midrule 
     UniMatch*~\cite{yang2023revisiting} \ \ \,| \textcolor{darkgray}{\textbf{RN-101 }} & \underline{73.1}	& \textbf{75.4}	& \underline{76.4}	& \underline{78.5}	& {79.3} \\
     {IPixMatch (ours)} | \textcolor{darkgray}{\textbf{RN-101 }} & \textbf{73.9} & \underline{74.6} & \textbf{77.1} & \textbf{78.9} & {79.4} \\ 
      \textit{Improvement} & \textit{+0.8} &	\textit{-0.7}  & \textit{+0.7}  & \textit{+0.4} & \textit{+0.1}  \\     
\midrule 
\bottomrule 
\end{tabularx}
\end{center}
\end{table*}

Given that UniMatch~\cite{yang2023revisiting} represents the latest algorithm that achieves state-of-the-art performance in  semi-supervised segmentation for both PASCAL VOC 2012 and Cityscapes datasets, we adopt the UniMatch as our primary baseline. Subsequently, we apply the proposed IPixMatch framework to UniMatch to further enhance its performance. 

The majority of our experimental settings mirror those of  UniMatch~\cite{yang2023revisiting}. For instance, we employ DeepLabV3+~\cite{chen2017deeplab} architecture with ResNet~\cite{he2016deep} (specifically, ResNet 50 and ResNet 101) as our segmentation backbone, paralleling the choices made in UniMatch. Of particular significance, we follow the exact same dataset split as in UniMatch, which minimizes randomness and ensures a fair basis for comparison. There are only two key differences. Firstly, due to the limited GPU resources, our overall batch size is halved. Accordingly, the learning rate is also reduced to half. Secondly, the three following configurations are implemented to guarantee fair comparison and precise reproduction of reported outcomes: (1) all IPixMatch experiments are implemented using PyTorch and are assigned with a consistent random seed (12345); (2) we configure PyTorch to eliminate any nondeterminism by setting  \textit{torch.use\_deterministic\_algorithms()} to \textit{True}; (3) in DeepLabV3+, multiple upsampling operations are present and often executed through bilinear interpolation with aligned corners. However, such operation can produce nondeterministic results. In our implementation, we opt for nearest interpolation instead to ensure reproduction. Note that segmentation performance is generally degraded after replacing bilinear interpolation with nearest interpolation~\cite{ahn2023comparison}. 

Since UniMatch is proved to outperform most state-of-the-art semi-supervised segmentation algorithms, we seek to validate the effectiveness of our proposed solution, IPixMatch, primarily through a comparison with UniMatch. To ensure a fair comparison between IPixMatch and UniMatch, we re-implement UniMatch by following the above-mentioned settings, which is denoted as UniMatch* thereafter. Other configurations are kept the same as UniMatch, please refer to~\cite{yang2023revisiting} for more details.

\subsection{Ablation study}\label{sec:ablation}

\begin{table*}[t!]
\caption{Comparison with SOTA methods on Cityscapes. The fractions in the head represent the proportion of labeled data. UniMatch* denotes our reproduced results of UniMatch, following the modifications mentioned in Sec.~\ref{sec: exp.detail}. Bold indicates the best, and italic shows the mIOU improvement of IPixMatch over UniMatch.}
\label{table:cityscape}
  \begin{center}
  \newcolumntype{Y}{>{\raggedleft\arraybackslash}X}
    \begin{tabularx}{\textwidth}{l|XXXX} 
     \toprule
      { \textbf{Cityscapes} } & 1/16 & 1/8 & 1/4 & 1/2 \\
      \midrule 
      Supervised Baseline & 63.3 & 70.2 & 73.1 & 76.6 \\
      PS-MT~\cite{liu2022perturbed} & - & 75.8 & 76.9 & 77.6 \\
      U$^2$PL~\cite{wang2022semi}  & 70.6 & 73.0 & 76.3 & 77.2 \\
      \midrule 

      UniMatch*~\cite{yang2023revisiting} \ \ \,| \textcolor{darkgray}{\textbf{RN-50 }} & {74.0}	& {75.0}	& {77.0}	& {77.5} \\
      {IPixMatch (ours)} | \textcolor{darkgray}{\textbf{RN-50 }}  & \textbf{74.1} &	\textbf{76.0}  & \textbf{77.5}  & \textbf{77.8} \\   
      \textit{Improvement} & \textit{+0.1} &	\textit{+1.0}  & \textit{+0.5}  & \textit{+0.3}  \\      
      \midrule 
      \midrule 
      Supervised Baseline & 66.3 & 72.8 & 75.0 & 78.0 \\
      CPS~\cite{chen2021semi} & 69.8 & 74.3 & 74.6 & 76.8 \\
      AEL~\cite{hu2021semi}& 74.5 & 75.6 & 77.5 & 79.0 \\
      PS-MT~\cite{liu2022perturbed} & - & 76.9 & 77.6 & \textbf{79.1} \\
     \midrule 
     UniMatch*~\cite{yang2023revisiting} \ \ \,| \textcolor{darkgray}{\textbf{RN-101 }} & {75.0}	&  {77.1}	& {78.0}	& {78.3}	\\
     {IPixMatch (ours)} | \textcolor{darkgray}{\textbf{RN-101 }} & \textbf{75.8} & \textbf{77.6} & \textbf{78.3} & {78.6} \\ 
      \textit{Improvement} & \textit{+0.8} &	\textit{+0.5}  & \textit{+0.3}  & \textit{+0.3} \\     
\midrule 
\bottomrule 
\end{tabularx}

\end{center}
\end{table*}

\textbf{Performance with different inter-pixel relation loss terms.}
In Sec.~\ref{sec: method-darkmatch}, two loss terms, namely $\mathcal{L}_{KL}$ and $\mathcal{L}_{CR}$, are introduced to mine inter-pixel relation with different distance metrics. However, they don't contribute equally to boost semi-supervised segmentation. By applying these terms above UniMatch, we individually assess their efficacy under the two protocols of PASCAL VOC 2012, as shown in Table~\ref{table:loss term} . Note that here we treat UniMatch as our baseline for these evaluations, and its re-implementation follows the same as in Sec.~\ref{sec: exp.detail} for fair comparisons. The balancing weight $\alpha$ in Eq.~\ref{eq:final_loss} is set to 1 for both cases.

From Table~\ref{table:loss term}, it can be observed that using KL distance is more beneficial to capture inter-pixel relation. One possible explanation for the inferior results of $\mathcal{L}_{CR}$ is that the output distribution difference between the teacher and student models in the semi-supervised framework is not severe enough. Although it is still difficult to measure quantitatively when correlation distance should be adopted instead of KL, we empirically assume that KL tends to be a more suitable choice for enforcing consistency when the teacher and student models share the same architecture, typically the scenario in semi-supervised learning.
Therefore, fore the rest of the paper, we use $\mathcal{L}_{KL}$ as the loss term to enforce inter-pixel consistency.

\textbf{Effect of warmup in mining of inter-pixel relation.}
Inter-pixel relation reveals additional information for semi-supervised segmentation. Nevertheless, in the early stage of training, the inter-pixel relation is more prone to be incorrect, potentially leading to misleading pseudo supervision. To counteract this challenge, we employ a straightforward strategy known as warmup to gradually increase the balancing weight $\alpha$ in Eq.~\ref{eq:final_loss} at the first few epochs. Specifically, following a prior work~\cite{laine2017temporal}, we adopt a Gaussian function $e^{-5(1-i_c/i_{w})^2}$ to ramp up $\alpha$, where $i_c$ and $i_w$ are the current iteration and the overall iterations in warmup, respectively. The outcomes of incorporating warmup, as presented in Table~\ref{table:warmup}, highlight improvements in segmentation performance. For our experiments conducted on PASCAL VOC 2012 and Cityscapes, the epochs for the warmup period are set as 5 and 30, respectively.

\subsection{Experimental results}
\textbf{PASCAL VOC 2012.} First, we compare our IPixMatch method with several recent state-of-the-art (SOTA) semi-supervised segmentation methods on PASCAL VOC 2012. The results are presented under two distinct protocols, \#P1 and \#P2, as shown in Table~\ref{table:pascal-P1} and Table~\ref{table:pascal-P2}, respectively. It is observed from Table~\ref{table:pascal-P1} that IPixMatch manages to consistently outperform other SOTAs regardless of the image resolution (321 or 513), encoder backbone (RN-50 or RN-101), and the split partitions (1/16, 1/8, or 1/4). In particular, the absolute improvements over the recent algorithm UniMatch are included for all settings (shown in italics). In general, IPixMatch exhibits more substantial performance improvements at lower input resolutions. Furthermore, the advantages of IPixMatch over UniMatch become more evident when fewer label data are utilized.

Table~\ref{table:pascal-P2} reports the results under split protocol \#P2. In comparison with the SOTA methods, IPixMatch achieves the highest mIOU in 7 out of 10 settings. 
In most cases, IPixMatch also outperforms UniMatch, with the exact gains quantified within the table. These improvements appears less pronounced as the volume of labeled data increases, suggesting diminishing returns in richer annotation contexts. Intriguingly, unlike the evaluation under protocols \#P1, the performance gain here does not exhibit consistent patterns, which could be related to the intrinsic complexity and diversity of the data samples in these splits. The exploration of the underlying causes for this variation is earmarked for future investigation.  

\textbf{Cityscapes.} In Table~\ref{table:cityscape}, the comparison on the more challenging  Cityscapes dataset is presented. With the backbone RN-50, IPixMatch achieves the highest mIoU across all partitions. Particularly, its performance surpasses UniMatch by 1.0\% under the 1/8 partition protocol. With the backbone RN-101, IPixMatch obtains the best performance when fewer labeled images are available (1/16, 1/8, and 1/4). When half of dataset is labeled, it shows slightly worse result (78.6\% vs 79.1\%), indicating that with more labelled data the structural consistency in urban scene of Cityscapes provides less scope for the inter-pixel relation enhancements. Nevertheless, IPixMatch consistently outperforms UniMatch across varying partitions, with the gain tapering as the number of labeled data increases.

\subsection{Qualitative results}
\begin{figure*}[t]
  \centering
   \includegraphics[width=1.0\linewidth]{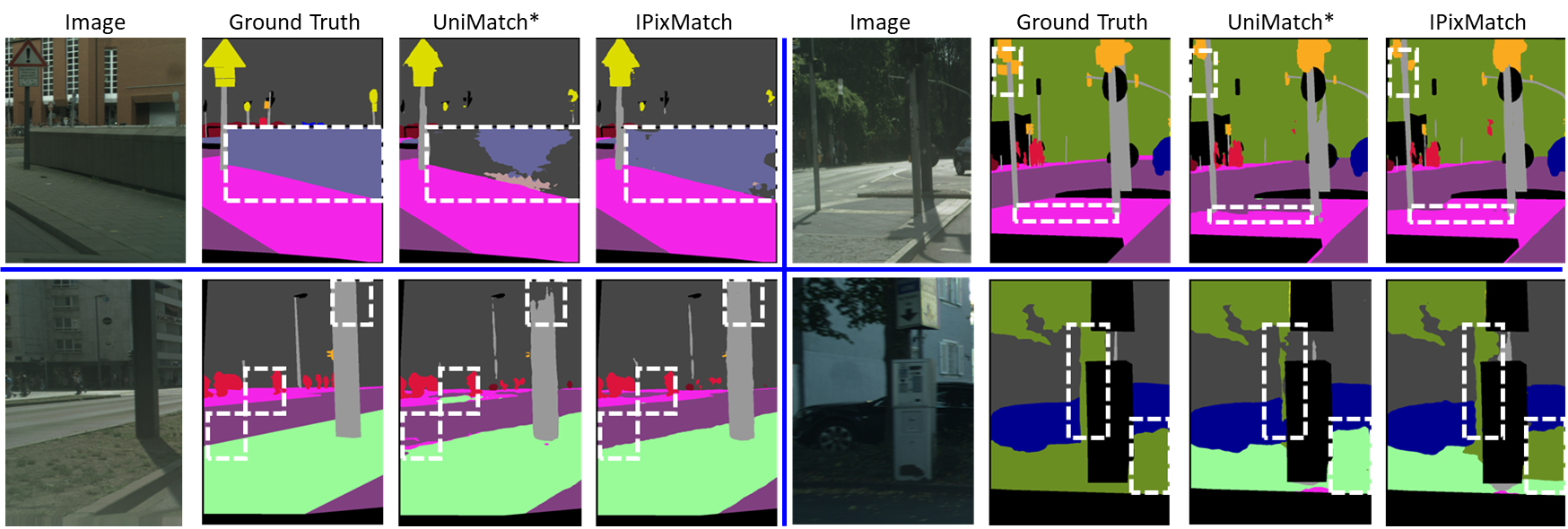}
   \caption{Qualitative comparison between IPixMatch and UniMatch* on Cityscapes: 1/8 labeled data with RN-50 as backbone. White dashed rectangles highlight the segmentation differences.}
   \label{fig:visulization}
\end{figure*}

The visualization in Fig.~\ref{fig:visulization} presents representative instances that highlight IPixMatch's superior performance compared to UniMatch within the same configuration. With the inter-pixel relation considered in IPixMatch, it allows the segmentation results to be more consistent with its surroundings.

\section{Conclusion}

In this work, we propose IPixMatch for enhancing semi-supervised semantic segmentation learning. Our motivation stems from the observations  that prior works often neglect the significance of inter-pixel relation, which carry essential information for accurate segmentation. To address this gap, we presented two loss terms that facilitate the exploration of inter-pixel relation within the framework of semi-supervised learning. IPixMatch integrates seamlessly into existing models without necessitating modifications or additional model blocks. Despite its simplicity, IPixMatch yields substantial improvements across many benchmarks under various partition protocols.

We envision the IPixMatch algorithm will stimulate heightened interest in the role of inter-pixel relation under the context of semi-supervised learning. Future work will explore whether such a solution can be extended to (1) other prediction tasks, such as instance segmentation, depth estimation, and panoptic segmentation, and (2) other scenarios involving learning with limited labeled data, such as domain adaptation. 






\bibliographystyle{IEEEtran}
\bibliography{ijcnn.bib}

\end{document}